\documentclass[letterpaper, 10 pt, conference]{ieeeconf}

\usepackage{amsthm}
\usepackage[english]{babel}

\newtheorem{remark}{Remark}
\IEEEoverridecommandlockouts
\usepackage{cite}

\usepackage{amsmath,amssymb,amsfonts}
\usepackage{algorithmic}
\usepackage{graphicx}
\usepackage{textcomp}
\usepackage{xcolor}
\usepackage[caption=false]{subfig}
\usepackage{multirow}
\usepackage{booktabs}
\usepackage[colorlinks,linkcolor=blue]{hyperref}
\usepackage{siunitx}
\usepackage{stfloats}
\usepackage{threeparttable}
\usepackage{algorithm}

\newcommand{\T}{\mathsf{T}}

\newcommand{\RR}{\mathbb{R}}

\DeclareMathOperator*{\argmax}{arg\,max}
\DeclareMathOperator*{\argmin}{arg\,min}

\newtheorem{lemma}{\bf Lemma}

\newtheorem{theorem}{\bf Theorem}
\newtheorem{proposition}{\bf Proposition}
\def\BibTeX{{\rm B\kern-.05em{\sc i\kern-.025em b}\kern-.08em
    T\kern-.1667em\lower.7ex\hbox{E}\kern-.125emX}}
\begin{document}

\title{\LARGE \bf Predictive Lagrangian Optimization for Constrained Reinforcement Learning}

\author{Tianqi Zhang$^{1\#}$, Puzhen Yuan$^{2\#}$, Guojian Zhan$^{3\#}$, Ziyu Lin$^{3}$, Yao Lyu$^{3}$, Zhenzhi Qin$^{5}$, \\
Jingliang Duan$^{6}$, Liping Zhang$^{5}$ and Shengbo Eben Li$^{3,4*}$
\thanks{*This work is supported by NSF China with U20A20334 and Tsinghua University Initiative Scientific Research Program. It is also supported by Tsinghua University-Toyota Joint Research Center for AI Technology of Automated Vehicle. T. Zhang, P. Yuan and G. Zhan contributed equally. All correspondences should be sent to S. E. Li with email: {\tt\small lishbo@tsinghua.edu.cn}.}
\thanks{$^{1}$Weiyang College, Tsinghua University, Beijing, 100084, China. $^{2}$Xingjian College, Tsinghua University, Beijing, 100084, China. $^{3}$School of Vehicle and Mobility, Tsinghua University, Beijing, 100084, China. $^{4}$College of Artificial Intelligence, Tsinghua University, Beijing, 100084, China. $^{5}$Department of Mathematical Sciences, Tsinghua University, Beijing, 100084. $^{6}$School of Mechanical Engineering, University of Science and Technology Beijing, Beijing, 100083.}%
}

\maketitle

\begin{abstract}

Constrained optimization is popularly seen in reinforcement learning (RL) for addressing complex control tasks.
From the perspective of dynamic system, iteratively solving a constrained optimization problem can be framed as the temporal evolution of a feedback control system.
Classical constrained optimization methods, such as penalty and Lagrangian approaches, inherently use proportional and integral feedback controllers.
In this paper, we propose a more generic equivalence framework to build the connection between constrained optimization and feedback control system, for the purpose of developing more effective constrained RL algorithms.
Firstly, we define that each step of the system evolution determines the Lagrange multiplier by solving a multiplier feedback optimal control problem (MFOCP). In this problem, the control input is multiplier, the state is policy parameters, the dynamics is described by policy gradient descent, and the objective is to minimize constraint violations.
Then, we introduce a multiplier guided policy learning (MGPL) module to perform policy parameters updating. And we prove that the resulting optimal policy, achieved through alternating MFOCP and MGPL, aligns with the solution of the primal constrained RL problem, thereby establishing our equivalence framework.
Furthermore, we point out that the existing PID Lagrangian is merely one special case within our framework that utilizes a PID controller. We also accommodate the integration of other various feedback controllers, thereby facilitating the development of new algorithms.
As a representative, we employ model predictive control (MPC) as the feedback controller and consequently propose a new algorithm called predictive Lagrangian optimization (PLO). 
Numerical experiments demonstrate its superiority over the PID Lagrangian method, achieving a larger feasible region up to $7.2\%$ and a comparable average reward.
\end{abstract}


\section{Introduction}

Reinforcement learning (RL) has shown remarkable potential across various domains, including video games\cite{mnih2015human}, Chinese Go\cite{Silver2016}, and robotics\cite{duan2021distributional}. However, real-world industrial control tasks, such as autonomous driving, demand strict safety constraint satisfaction \cite{zhan2024transformation} and remain a vital challenge for existing RL algorithms \cite{li2023rlbook, guan2021direct}. 

To tackle constraints, researchers have integrated various constrained optimization techniques into standard RL algorithms, which are also known as constrained RL algorithms. In the early stages, penalty methods\cite{yoo2021dynamic}, such as interior-point and exterior-point, were widely adopted due to their ease of implementation\cite{ren2022self}. However, adding a penalty term to the objective with a fixed ratio will inevitably lead to an optimal solution drift. Therefore, recent research has majorly turned to the Lagrange multiplier method\cite{paternain2019learning}.
For each constraint term, this approach leverages a scalar called Lagrange multiplier, to integrate it into the primal objective and finally obtain a scalar criterion called Lagrangian. By further employing dual-descent-ascent training technique, this method explores the solution space of both policy parameters and multiplier to strike a balance between maximizing performance and satisfying constraints. Nevertheless, properly managing the multiplier update remains a challenge in this minimax training paradigm and is prone to causing significant oscillations.

Recently, a dynamic system perspective on optimization problems has emerged, offering the potential for designing more effective algorithms. Michael Jordan et al. were among the first to propose that iteratively solving an optimization problem can be conceptualized as the temporal evolution of a dynamic system\cite{jordan2018dynamical}. As the dynamic system autonomously evolves, its state gradually converges to the equilibrium, which is analogous to the convergence of variables in an optimization problem toward the optimal solution.
When it comes to solving the constrained optimization problems, Stooke et al. revealed that it can be framed as the temporal evolution of a feedback control system, where the constrained violation and multiplier are feedback error and control input, respectively\cite{pmlr-v119-stooke20a}. 
In this context, classical constrained optimization methods such as penalty and Lagrangian methods inherently use proportional and integral controllers.
PID Lagrangian method, as a rainbow-like synthesizer, employs a PID controller and achieves empirically promising performance.
Subsequently, several variations by modifying the PID controllers are proposed with satisfactory performance and near-zero constraint violations \cite{peng2021separated, peng2021model}.
Intuitively, this feedback control mechanism can be interpreted as controlling multiplier to minimize policy's constraint violations while satisfying the dynamics represented by policy gradient descent, thereby essentially addressing another problem. 
However, the existing literature lacks a comprehensive investigation into the relationship between the essentially solving problem and the primal constrained RL problem, leading to limited interpretability.
Meanwhile, as PID is the most basic form of feedback control, an exciting opportunity arises:  what if employing other well-established control methods to serve as the feedback controller?

In this paper, we investigate the connection between constrained optimization and feedback control systems, and firstly establish a generic framework to build their equivalence. Our framework consists of two alternating modules: a multiplier feedback optimal control problem (MFOCP) module for multiplier updating, and a multiplier-guided policy learning (MGPL) module for policy updating. Notably, the PID Lagrangian method is just a special case within our framework that uses a PID controller. And we allow the incorporation of any other feedback controllers for multiplier updating to design new algorithms. On this basis, we propose a new constrained RL algorithm called predictive Lagrangian optimization (PLO), which leverages model predictive control (MPC) to serve as the feedback controller for further performance enhancement. 
The key contributions of this paper are summarized as: 

\begin{enumerate}
    \item We establish a generic equivalence framework to connect the iterative solving process of a constrained optimization problem with the temporal evolution of a feedback control system. We achieve this through two key procedures. Firstly, we define that each step of the system evolution determines the Lagrange multiplier by solving a multiplier feedback optimal control problem (MFOCP), where the control input is multiplier, the state is policy parameters, the dynamic is described by policy gradient descent, and the objective is to minimize a feedback error that reflects constraint violations. Then, we introduce a multiplier-guided policy learning (MGPL) module to perform policy parameter updating. And we prove that the resulting optimal policy by alternating MFOCP and MGPL, aligns with the solution of the primal constrained RL problem, thereby establishing their equivalence.
    This framework opens a new avenue for leveraging various advanced controllers for multiplier updating, thereby facilitating the design of more effective constrained RL algorithms.
    \item As a representative, we propose a constrained RL algorithm called PLO, which employs MPC as the feedback controller. The feedback error (i.e., the MFOCP objective) is designed as the cumulative constraint violations within a prediction horizon.
    Unlike PID Lagrangian which considers only the current constraint violation, PLO enjoys the receding horizon capabilities of MPC to take predicted constraint violations into consideration. Consequently, our PLO demonstrates better training efficiency and excels in prioritizing safety while maintaining high policy performance.

\end{enumerate}

Numerical experiments on classical control tasks demonstrate that our PLO achieves a higher level of safety by expanding the feasible region up to 7.2\% and has a comparable average reward compared with PID Lagrangian.


\section{Preliminaries}
\label{sec:preliminaries}
This section introduces the principles of constrained RL, and elaborate on MPC utilized in our algorithm design.

\subsection{Constrained Reinforcement Learning}

Constrained RL extends standard RL by incorporating constraints that must be satisfied while learning optimal policies.
The general formulation of constrained RL is 

\begin{equation}
\begin{aligned}
\max_\theta J(\theta)&=\mathbb{E}\bigg\{\sum_{i=0}^{\infty} \gamma^i r(x_i, u_i)\bigg\},\\
\quad \mathrm{s.t.}\quad
x_{i+1}&=f(x_{i},u_i),\\
h(s_{i})&\leq 0,i=0,1,\cdots,\infty,
\label{eq.constrained_rl}
\end{aligned}
\end{equation}
where the state $x\in \RR^n$, action $u\in\RR^m$, dynamics $f:\RR^n\times  \RR^m  \mapsto \RR^n$ and the policy is parameterized by $\theta$. The reward and cost signals are represented by $r\in \RR$ and $h\in \RR$.
\begin{remark}
    If we define that the cost signal is non-positive, i.e., $c(x) = \max\{h(x), 0\} \ge 0$, the formulation \eqref{eq.constrained_rl} also can be expressed as
\begin{equation}
\begin{aligned}
\label{eq:original form}
    &\max_\theta{ J(\theta) } \\
    \text{s.t}&. \ J_c(\theta) \leq 0,
\end{aligned}
\end{equation}
where $J_c(\theta) = \mathbb{E}\big\{ \sum_{i=0}^{\infty} \gamma^i c(x_{i}) \big\}$.
\end{remark}

\subsection{Model Predictive Control} 
MPC is an advanced feedback controller that leverages an analytical dynamic model to conduct receding horizon control. Its objective function is formulated as
\begin{align}
    J_{\text{MPC}} = \sum_{i=0}^{N-1} (x_{i \vert k} - x^\text{ref}_{i \vert k})^\T Q (x_{i \vert k} - x^\text{ref}_{i \vert k}) + u_{i \vert k}^\T R u_{i \vert k},
\end{align}
where $N$ is the length of prediction horizon. The footnote $(i \vert k)$ indicates the $i$-th moment in the prediction horizon starting from time $k$ in the real-time domain. The total objective $J_{\text{MPC}}$ is the accumulated errors and action consumption over the prediction horizon.
MPC solves an optimal action sequence, denoted as $u_{0 \vert k}^*, u_{1 \vert k}^*, ..., u_{N-1 \vert k}^*$, and implement the first one in the real-time domain at each step, which is known as receding horizon control. 

\section{Method}
\label{sec:method}

In this section, we first present the established generic equivalence framework to connect constrained optimization and feedback control systems. Then we prove the optimality of the resulting policy for the primal constrained RL problem.
Finally, we elaborate on our PLO algorithm.

\subsection{Framework for Connecting Constrained Optimization and Feedback Control System}

\begin{figure*}[ht]
\centering
\includegraphics[width=0.95\linewidth,trim={0.5cm 1.0cm 0.5cm 0.7cm}, clip]{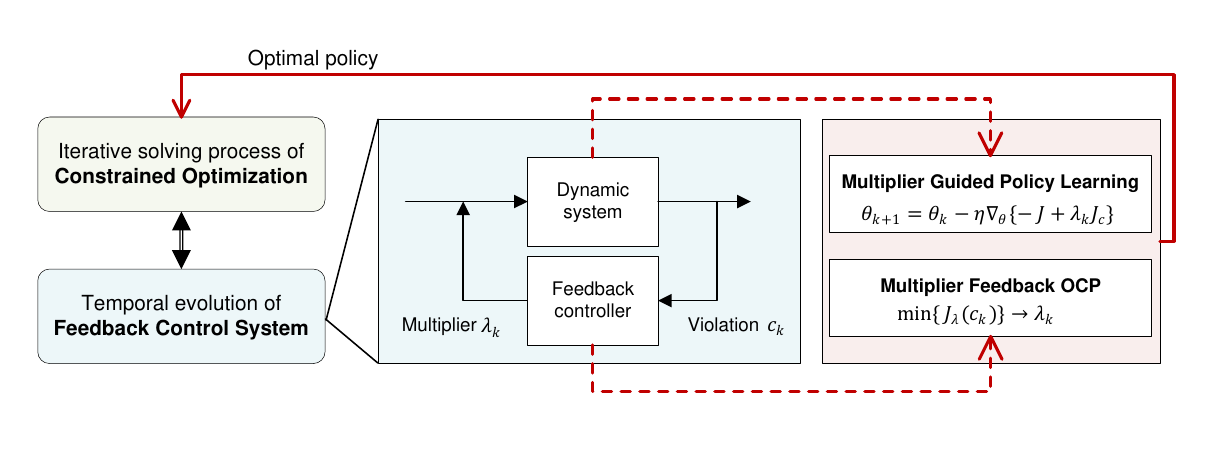}
\caption{Framework for connecting constrained optimization and feedback control systems.}
\label{fig.framework}
\end{figure*}


We begin with applying dual transformation on the primal constrained optimization problem \eqref{eq:original form}, resulting in
\begin{equation}
\begin{aligned}
\label{eq:dual form}
    \underset{\lambda \ge 0}{\max} 
    \underset{\theta}{\min}
    -J(\theta) + \lambda J_c(\theta).
\end{aligned}
\end{equation}
To solve this two-loop minimax problem, our framework, as shown in Figure \ref{fig.framework}, consists of two alternating modules: an MFOCP module for updating the outer loop $\lambda$ and an MGPL module for updating the inner loop $\theta$ guided by $\lambda$.

We first introduce the formulation of MFOCP, aiming to address the outer loop of \eqref{eq:dual form}. According to the perspective of feedback control dynamic system and use $k$ to denote the iteration index, 
we perceive the policy parameter $\theta_k$ as the system state, the Lagrange multiplier $\lambda_k$ as the control input and the constraint violation $J_c({\theta_k})$ as the feedback error. A straightforward choice is to employ a feedback controller to adjust the multiplier, such as
$    \lambda_k = \lambda_{k-1} + \eta J_c({\theta_k})$, to balance performance and safety, where $\eta$ is a learning rate. However, this approach is somewhat intuitive and there is significant room for improvement.
Here, we formulate an optimal control problem form, termed MFOCP, to solve an optimal multiplier at each iteration, which is given by
\begin{equation}
\begin{aligned}
\label{eq:PID target}
    \underset{\lambda \ge 0}{\min} 
    |J_c({\theta(\lambda)})|& \\
    \text{s.t.} \ \theta = \theta(\lambda) \in \underset{\theta}{\argmin} -J(&\theta) + \lambda J_c(\theta),
\end{aligned}
\end{equation}
where $\theta(\lambda)$ implies that $\theta$ is an implicit function of $\lambda$.  The significance of the MFOCP module is that any feedback controller capable of solving \eqref{eq:PID target}  can be incorporated to update the multiplier, consequently making up a new algorithm.

Next, for the inner loop of \eqref{eq:dual form}, given the solved multiplier $\lambda_k$ at each iteration $k$, the MGPL module leverages this current step multiplier to build the current step policy training objective and update the policy as
\begin{equation}
\label{eq:MGPL}
    \theta_{k+1} = \theta_{k} - \eta (-\nabla_\theta J({\theta_{k}}) + \lambda_{k} \nabla_\theta J_c({\theta_{k}}))
\end{equation}


\subsection{Optimality Analysis of Multiplier Feedback Updating}

This subsection demonstrates the equivalence of solving the primal problem \eqref{eq:original form}, its dual problem \eqref{eq:dual form}, and the MFOCP \eqref{eq:PID target}, i.e., establishing the optimality of our framework.
Specifically, we complete this proof through three procedures:
\begin{enumerate}
    \item Prove the monotonicity of $J_c({\theta(\lambda)})$ in Proposition \ref{prop:monotonicity}.
    \item Demonstrate an equality property of monotonic functions in Lemma \ref{prop:math relation}.
    \item Utilize Proposition \ref{prop:monotonicity} and Lemma \ref{prop:math relation} to prove Theorem \ref{prop:equivalency},  establishing the optimality.
\end{enumerate}

We begin with the monotonicity proposition as follows.
\begin{proposition}
\label{prop:monotonicity}
Assume $-J(\theta)$ and $J_c(\theta)$ are differentiable and strongly convex with respect to $\theta$, then $J_c({\theta(\lambda)})$ is a strictly monotonically decreasing function of $\lambda \in [0, +\infty)$.
\end{proposition}

\begin{proof}

Given the differentiability and strong convexity of $-J(\theta)$ and $J_c(\theta)$, $\theta = \theta(\lambda)$ is uniquely determined and can be represented by a differential relation as follows:
\begin{equation}
\begin{aligned}
\label{eq:differential relation 1}
    \theta = \theta(\lambda) \iff
    -\nabla_\theta J(\theta) + \lambda \nabla_\theta J_c(\theta) = 0.
\end{aligned}
\end{equation}

By taking the derivative of both sides of \eqref{eq:differential relation 1} and using the implicit function theorem, we have
\begin{equation}
\begin{aligned}
\label{eq:differential relation 2}
    \frac{\mathrm{d} \theta}{\mathrm{d} \lambda} =
    -\nabla_\theta^2 L(\theta, \lambda)^{-1} 
    \nabla_\theta J_c(\theta) |_{\theta=\theta(\lambda)},
\end{aligned}
\end{equation}
where
$
L(\theta, \lambda)=-J(\theta) + \lambda J_c(\theta)
$. We further considering the derivative of $J_c({\theta(\lambda)})$ with respect to $\lambda$ and obtain
\begin{equation}
\begin{aligned}
\label{eq:differential relation 3}
    \frac{\mathrm{d}}{\mathrm{d} \lambda}
    J_c({\theta(\lambda)}) = 
    \nabla_\theta J_c(\theta)^\T
    \frac{\mathrm{d} \theta}{\mathrm{d} \lambda} \Big|_{\theta=\theta(\lambda)}.
\end{aligned}
\end{equation}

By substituting \eqref{eq:differential relation 2} into \eqref{eq:differential relation 3}, we have
\begin{equation}
\begin{aligned}
\label{eq:differential relation 4}
    \frac{\mathrm{d}}{\mathrm{d} \lambda}
    J_c({\theta(\lambda)})=-\nabla_\theta J_c(\theta)^\T
    \nabla_\theta^2 L(\theta, \lambda)^{-1} 
    \nabla_\theta J_c(\theta) |_{\theta=\theta(\lambda)}.
\end{aligned}
\end{equation}

Notice that we assume $-J(\theta)$ and $J_c(\theta)$ are both differentiable and strongly convex, which implies that the Hessian matrix $\nabla_\theta^2 L(\theta, \lambda)$ is positive definite. Consequently, we have 
\begin{equation}
\begin{aligned}
\label{eq:differential relation 5}
    \forall \lambda > 0, \
    \frac{\mathrm{d}}{\mathrm{d} \lambda}
    J_c({\theta(\lambda)}) < 0,
\end{aligned}
\end{equation}
which supports the strictly monotonically decrease property of $J_c({\theta(\lambda)})$.



\end{proof}

Then the subsequent lemma states an equality property of the monotonic function involving its value and derivative.

\begin{lemma}
\label{prop:math relation}
If $f(x) \in C^1(\mathbb{R}^+)$ and $f'(x)$ strictly monotonically decreases, then
\begin{equation}
\begin{aligned}
\label{eq:math relation}
    \underset{x \ge 0}{\argmax} f(x) =
    \underset{x \ge 0}{\argmin} |f'(x)|.
\end{aligned}
\end{equation} 
\end{lemma}

\begin{proof}

We first denote that $x^\ast \in \argmax_x f(x)$ and $x' \in \argmin_x |f'(x)|$. 
If $x^\ast=0$, then we have $f'(0) \le 0$. Considering $f'(x)$ is a strictly monotonically decreasing function, $\forall x \ge 0$, $f'(x)$ is negative and decreases, which implies that $|f'(x)|$ increases and $x^\ast \in \argmin_x |f'(x)|$. If $x^\ast>0$, considering $f(x) \in C^1(\mathbb{R}^+)$, $f'(x^\ast)=0$ must hold and thus $x^\ast \in \argmin_x |f'(x)|$. Therefore, $\argmax_x f(x) \subseteq \argmin_x |f'(x)|$. 

If $x'=0$, considering $f'(x)$ is a strictly monotonically decreasing function, then we have $f'(0) \le 0$. This inequality arises from the fact that if $f'(0)>0$, there exist points near $x'$ whose function values are closer to zero, leading to a contradiction. Consequently, $f'(x) \le 0 $ and it is obvious that $x' \in \argmax_x f(x)$. If $x'>0$, then we have $f'(x') = 0$. This is because if $f'(x') \not= 0$, there exist points near $x'$ whose function values are closer to zero, leading to a contradiction. As $\forall x < x', f'(x)>f'(x')=0$ and $\forall x > x', f'(x)<f'(x')=0$, we have $x' \in \argmax_x f(x)$. Therefore, $\argmin_x |f'(x)| \subseteq \argmax_x f(x)$. 

To sum up, we have $\argmax_x f(x) = \argmin_x |f'(x)|$. 

\end{proof}

After establishing the inherent monotonicity of $J_c({\theta(\lambda)})$ in Proposition \ref{prop:monotonicity} and demonstrating its relevant property in Lemma \ref{prop:math relation}, we present the following Theorem \ref{prop:equivalency} to establish the equivalence between solving \eqref{eq:original form} and  \eqref{eq:PID target}.

\begin{theorem}[Optimality]
\label{prop:equivalency}
The solution to the primal problem \eqref{eq:original form} aligns with that of \eqref{eq:PID target}, assuming that $J(\theta)$ and $J_c(\theta)$ are differentiable and strongly convex.
\end{theorem}

\begin{proof}

We first consider the equivalence between the dual problem \eqref{eq:dual form} and MFOCP \eqref{eq:PID target}. Given \eqref{eq:differential relation 1}, by denoting the dual objective as
$
\Gamma(\lambda) 
= \min_\theta L(\theta, \lambda)
= L(\theta(\lambda), \lambda)
$
and using the derivation rule for composite functions, we have
\begin{equation}
\begin{aligned} 
\label{eq:dual target relation}
    \Gamma(\lambda)'
    &= \nabla_\theta L(\theta, \lambda) ^\T \frac{\mathrm{d} \theta}{\mathrm{d} \lambda} \Big| _{\theta=\theta(\lambda)} +
    \frac{\partial L}{\partial \lambda} \Big|_{\theta=\theta(\lambda)} \\
    &= \frac{\partial L}{\partial \lambda} \Big|_{\theta=\theta(\lambda)}\\
    &=J_c({\theta(\lambda)}). \\
\end{aligned}
\end{equation}
Then we denote the solution of \eqref{eq:dual form} as $(\theta^\ast, \lambda^\ast)$, which satisfies
\begin{equation}
\begin{aligned}
\label{eq:dual solution}
    \lambda^\ast \in \underset{\lambda \ge 0}{\argmax}
    \Gamma(\lambda), \
    \theta^\ast = \theta(\lambda^\ast).
\end{aligned}
\end{equation}
Recall that Proposition \ref{prop:monotonicity} demonstrates that $\Gamma(\lambda)'=J_c({\theta(\lambda)})$ is a strictly monotonically decreasing function, and by further invoking Lemma \ref{prop:math relation}, we have
\begin{equation}
\begin{aligned} 
\label{eq:equivalency}
    \underset{\lambda \ge 0}{\argmax} \Gamma(\lambda) = 
    \underset{\lambda \ge 0}{\argmin}
    |J_c({\theta(\lambda)})|.
\end{aligned}
\end{equation}

Therefore, the solution of the dual problem \eqref{eq:dual form} is equivalent to that of MFOCP \eqref{eq:PID target}. Furthermore, given the differentiable and strong convexity assumption, the solution of dual problem \eqref{eq:dual form} is also equivalent to that of its primal problem \eqref{eq:original form}.
This completes the proof.


\end{proof}

\subsection{Algorithm Design}



Theorem \ref{prop:equivalency} allows us to leverage diverse well-established control approaches to serve as the multiplier feedback controller with theoretical soundness. Here, we employ MPC as the feedback controller and propose the PLO algorithm.




Specifically, to align the solution of MFOCP in \eqref{eq:PID target}, the formulation of MPC at each iteration $k$ is designed as
\begin{equation}
\begin{aligned}
\label{eq:plo}
    \min_{\lambda_{0\vert k}, \dots, \lambda_{N-1\vert k}} J_{\text{MPC}}&=\sum_{i=0}^{N-1} J_c({\theta_{i \vert k}})^2 + R \lambda_{i \vert k}^2,\\
    \text{s.t.} \ 
    \theta_{i+1|k} &= \theta_{i|k} + \eta(\nabla_\theta J({\theta_{i|k}}) - \lambda_{i|k} \nabla_\theta J_c({\theta_{i|k}})) \\
     \lambda_{i \vert k} &\geq 0, \ \forall i=0,\dots, N-1,
\end{aligned}
\end{equation}
where $N$ is the length of prediction horizon, $R$ serves as the multiplier  regularization weight. Algorithm \ref{MPC algo} demonstrates our PLO algorithm in detail.

\begin{algorithm}[H]
\caption{Predictive Lagrangian Optimization} 
\label{MPC algo}
\begin{algorithmic}
\STATE \textbf{Hyperparameters}: $N, R, \eta$
\STATE Initialize policy parameter $\theta_0$, iteration $k=0$
\REPEAT
    \STATE Calculate $\lambda_{0 \vert k}^*, \lambda_{1 \vert k}^*, ..., \lambda_{N-1 \vert k}^*$ by \eqref{eq:plo}\\
    \STATE Set $\lambda_k = \lambda_{0 \vert k}^*$ \\
    \STATE Update $\theta_{k+1} = \theta_k + \eta(\nabla_\theta J({\theta_k}) - \lambda_k \nabla_\theta J_c({\theta_k}))$
    \STATE $k = k + 1$
\UNTIL{convergence}


\end{algorithmic}
\end{algorithm}

\section{Experiment}
\label{sec:experiment}
In this section, we conduct numerical experiments on two classical control tasks involving a double integrator and a cartpole, to validate the efficacy of our PLO algorithm.

\subsection{Tasks}
\subsubsection{Double Integrator}

The dynamics is formulated as
\begin{equation}
\begin{aligned}
\label{eq:dynamic equation}
    \dot{x} = \mathbf{A} x + \mathbf{B} u,
\end{aligned}
\end{equation}
where $x = [x_1, x_2]^\T \in\mathbb{R}^2$, and $u\in\mathbb{R}$, and the matrices $A$ and $B$ are
\begin{equation}
\begin{aligned}
\label{eq:A_B_matrices}
    \mathbf{A} &= 
    \begin{bmatrix}
    0 & 1 \\
    0 & 0
    \end{bmatrix},
    \quad
    \mathbf{B} &= 
    \begin{bmatrix}
    0 \\
    1
    \end{bmatrix}.
\end{aligned}
\end{equation}
The objective is to drive all states toward zero, and the reward signal is designed as
\begin{equation}
\begin{aligned}
\label{eq:reward}
    r(x,u)=-x_1^2-x_2^2.
\end{aligned}
\end{equation}

Regarding the constraints, we ascertain that the system is safe when $x_1$ stays within $[1, 5]$, i.e., $h_1(x) = 1 - x_1 \leq 0$ and $h_2(x) = x_1 - 5 \leq 0$. 
Therefore, the cost signal is
\begin{equation}
c(x)=
\begin{cases}
1-x_1  &\text{if} \ x_1 < 1 \\
x_1 - 5  &\text{if} \ x_1 > 5 \\
0  &\text{else} \\
\end{cases}.
\end{equation}

\subsubsection{Cartpole}

The dynamic is formulated as
\begin{equation}
\begin{aligned}
\label{eq:inverted pendulum's dynamic equation}
    \ddot{p} = \frac{F + m \cdot l \cdot \dot{\varphi}^2 \cdot \sin(\varphi)}{M + m} - \frac{m \cdot l \cdot \ddot{\varphi} \cdot \cos(\varphi)}{M + m}, \\
    \ddot{\varphi} = \frac{g \cdot \sin(\varphi) - \cos(\varphi) \cdot \left(\frac{F + m \cdot l \cdot \dot{\varphi}^2 \cdot \sin(\varphi)}{M + m}\right)}{l \cdot \left(\frac{4}{3} - \frac{m \cdot \cos^2(\varphi)}{M + m}\right)},
\end{aligned}
\end{equation}
where the state $x=[p, \dot{p}, \varphi, \dot{\varphi}]^\T\in\mathbb{R}^4$, consisting of the position, velocity, angle, and angular velocity of the pole. The action $u=F\in\mathbb{R}$ is the horizontal force applied on the cart. The other dynamical variables including cart mass $M$, pole mass $m$, pole length $l$, and gravitational acceleration $g$ follow the standard settings of OpenAI Gym \cite{wang2023gops}.
The objective is to keep the pole vertically balanced, and the reward signal is designed as
\begin{equation}
    r(x,u)= - 10 \cdot \varphi^2,
\end{equation}
to encourage the pole to stand upright.

Regarding the constraint, we ascertain that the system is safe when $x$ stays within $[-1, 1]$, i.e., $h_1(x) = -1 - x \leq 0$ and $h_2(x) = x - 1 \leq 0$.
Therefore, the cost signal is
\begin{equation}
c(x)=
\begin{cases}
-1 - x  &\text{if} \ x < -1 \\
x - 1  &\text{if} \ x > 1 \\
0  &\text{else} \\
\end{cases}.
\end{equation}

\subsection{Settings}
There are some commonalities in those two experiments.
\subsubsection{Methods}
We employ the finite-horizon approximate dynamic programming (FHADP) implemented in GOPS as the backbone RL algorithm \cite{wang2023gops, zhan2023enhance}. Then we equip it with feedback controllers for multiplier updating to address constraints following our equivalence framework. Recall that PID controller leads to the PID Lagrangian method and MPC controller results in our proposed PLO method.
The policy function is parameterized by a three-layer multilayer perception, employing tanh activation function and consisting of 64 units in each hidden layer. Other hyperparameters can be found in Table \ref{tab:hyperparameter}.

\begin{table}[H]
  \begin{center}
    \caption{Hyperparameter for  experiment}
    \label{tab:hyperparameter}
    \begin{tabular}{l l l} 
    \hline
        \textbf{Symbol} & \textbf{Description} & \textbf{Value} \\ \hline
        $K_P$ & Coefficient of the proportional term & $1\times 10^{-2}$\\
        $K_I$ & Coefficient of the integral term & $1\times 10^{-4}$\\
        $K_D$ & Coefficient of the derivative term & $1\times 10^{-4}$\\
        $N_{f}$ & Prediction horizon in FHADP & 80\\
        $N$ & Prediction horizon of MPC in PLO & 20\\
        $R$ & Control regularization weight& $1\times 10^{-4}$\\
        \hline
    \end{tabular}
  \end{center}
\end{table}

\subsubsection{Evaluation}
For each policy checkpoint, we conduct the following evaluation: initial states are enumerated within the whole state space. From each initial state, we simulate forward 200 steps using the policy at that particular checkpoint. We record the maximum constraint violation and the mean reward along this simulated trajectory. Specifically, two cases are considered infeasible: (1) if the initial state has already violated the constraint, then we label it as infeasible point, i.e., \textit{initial infeasible}; (2) if the maximum constraint violation along the trajectory starting from the initial state exceeds a predefined safety threshold, i.e., 0.1, we label the initial state as infeasible point., i.e., \textit{endless infeasible} \cite{li2023rlbook}. 

\subsection{Results}
\begin{figure*}[h]%
    \centering
    \subfloat[Double integrator]{\includegraphics[width=0.45\linewidth,trim={0.8cm 0cm 1cm 1.5cm}, clip]{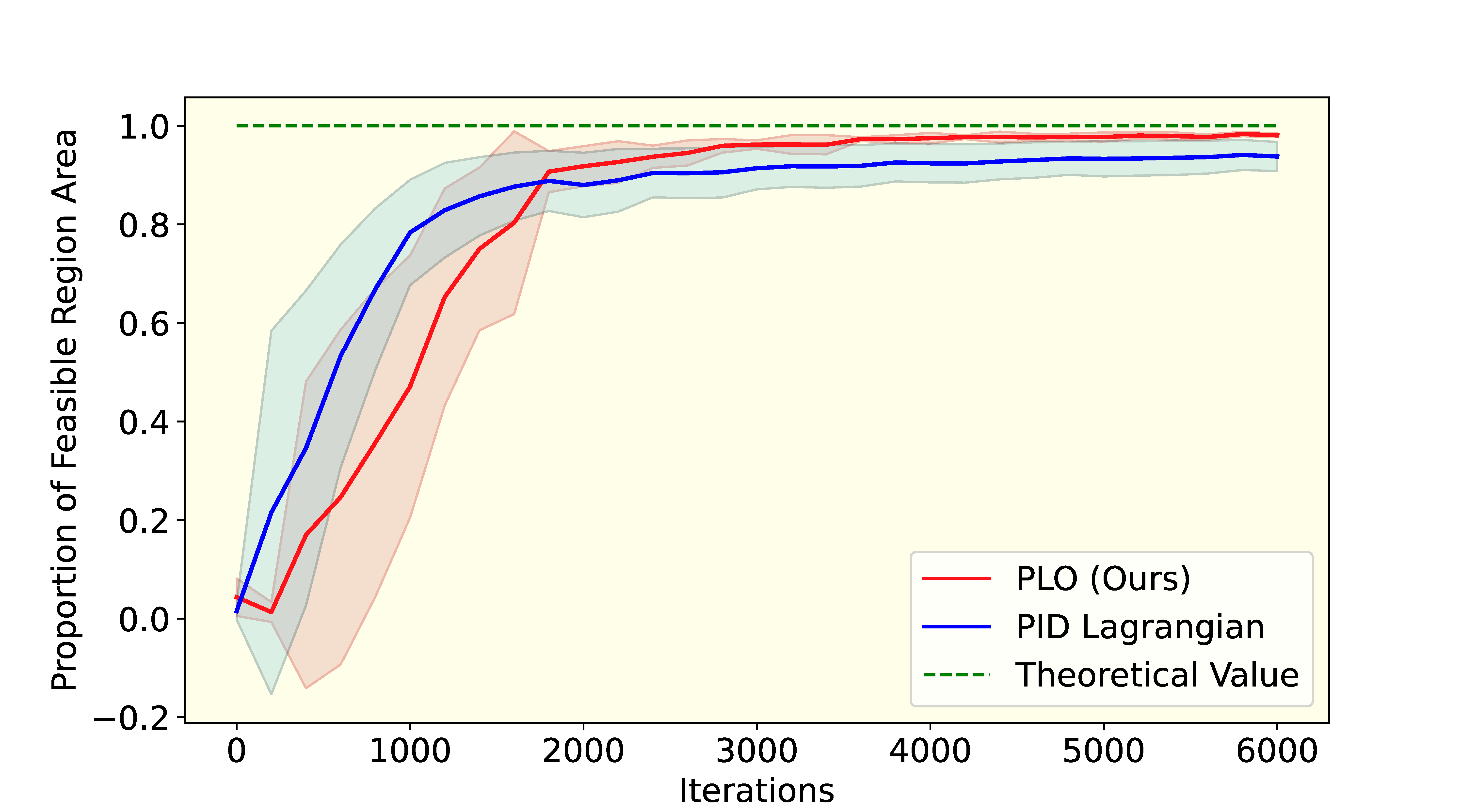}}
    \subfloat[Cartpole]{\includegraphics[width=0.45\linewidth,trim={0.8cm 0cm 1cm 1.5cm}, clip]{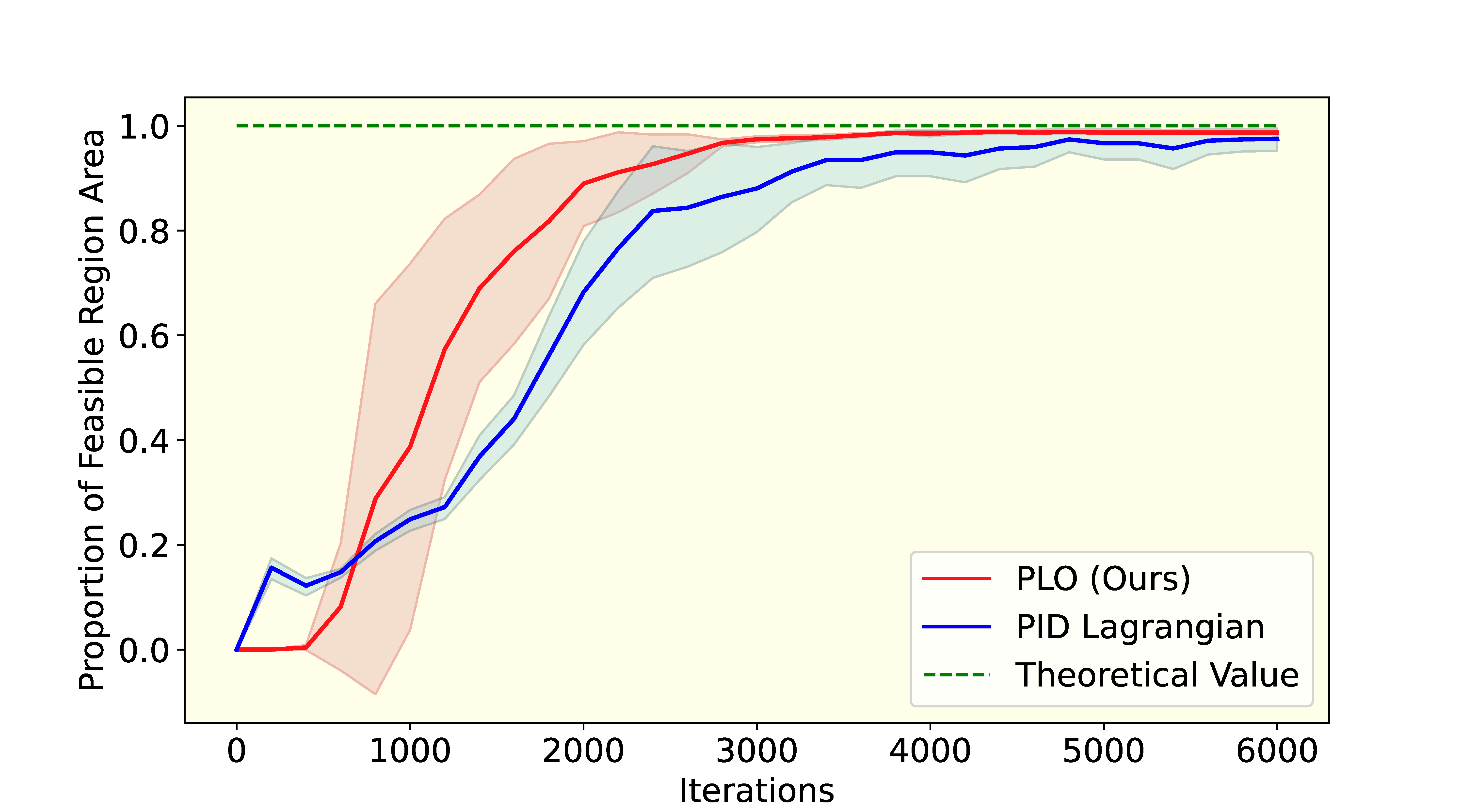}}
    \caption{Feasible region comparison. During the training process, we periodically save a policy per 200 iterations, identify and quantify its feasible region area. The x-axis is the number of training iterations, and the y-axis is the proportion between the feasible region area and its theoretical maximum value.}
    \label{region size}
\end{figure*}

\begin{figure*}[htbp!]
\centering
    \subfloat[Feasible region expanding process of PLO (red) and PID Lagrangian (blue).]{\includegraphics[width=0.98\linewidth]{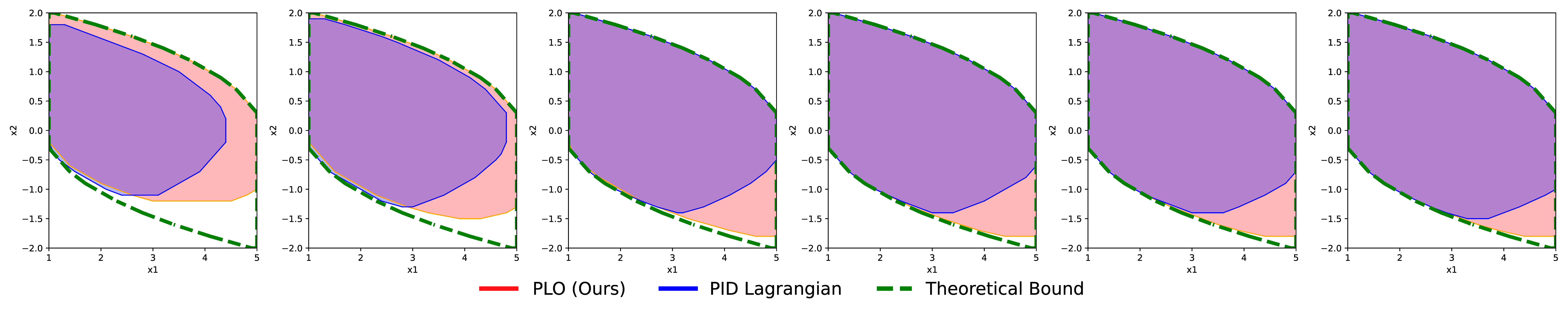}
    \label{subfig_fr}} \\
    \subfloat[Mean reward comparison of PLO (top) and PID Lagrangian (bottom).]{\includegraphics[width=0.98\linewidth,trim={0cm 8cm 0cm 7.5cm}, clip]{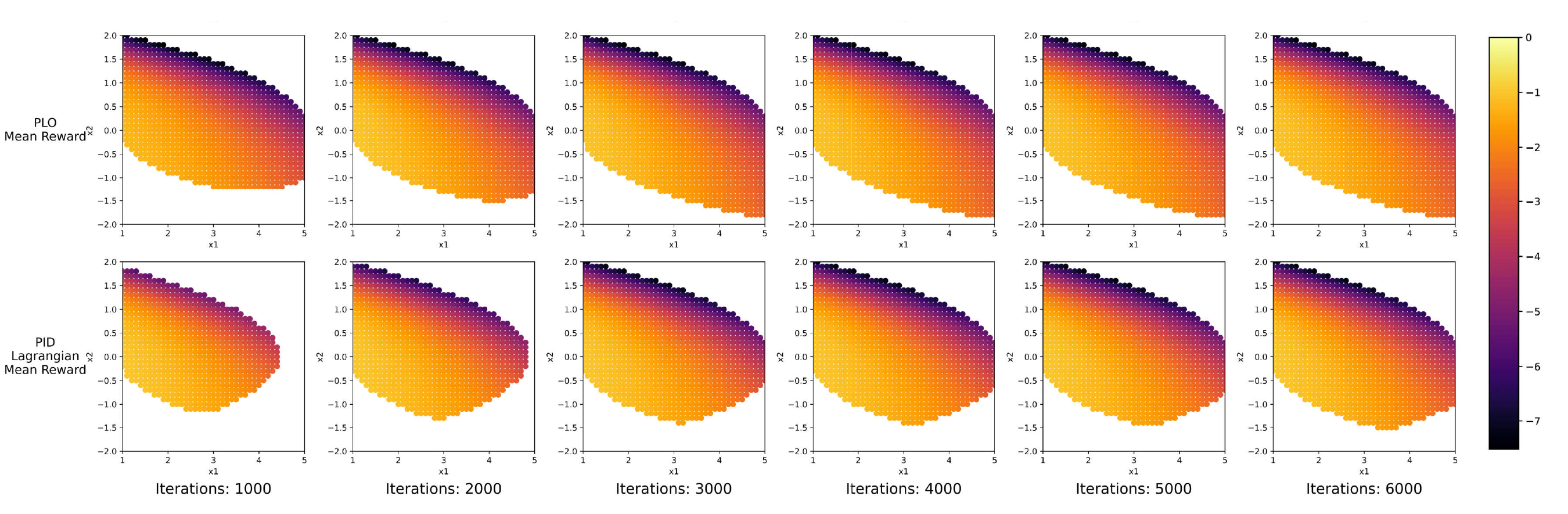}
    \label{subfig_rew}}
    \caption{Feasible region and mean reward visualization during training process in the double integrator experiment. Sub-figure (a) illustrates the feasible region, where the red and blue areas represent the feasible region obtained by our PLO algorithm and PID Lagrangian algorithm, respectively. The green dotted line outlines the theoretical maximum feasible region. Sub-figure (b) displays the mean reward along the evaluation trajectories, with lighter colors indicating higher values. In all plots, the x-axis denotes the initial value of $x_1$, ranging from 1 to 5, while the y-axis represents $x_2$ ranging from -2 to 2.}
    \label{eval results}
\end{figure*}

To quantify the safety, we compute the theoretical maximum feasible region and present the proportion between the obtained feasible region area and its theoretical maximum value in Fig. \ref{region size}. The results show that as training progresses, PLO's feasible region expands rapidly, and its curve consistently surpasses the curve corresponding to that of PID Lagrangian. Notably, at around 4,000 iterations, PLO's feasible region reaches approximately its theoretical maximum extent. Take the double integrator experiment as an instance, PLO exhibits a significant improvement in terms of safety, with up to 7.2\% larger feasible region than that obtained by PID Lagrangian at the end of training. Fig. \ref{subfig_fr} presents a specific visualization of the feasible region expanding process.
Additionally, the mean reward comparison shown in  Fig. \ref{subfig_rew} demonstrates that our PLO achieves comparable performance to the PID Lagrangian. Given that the former method has a larger feasible region compared to the latter, it is evident that our PLO effectively prioritizes safety while still maintaining high policy performance.

\section{Conclusion}
\label{sec:conclusion}
In this paper, we establish a generic equivalence framework to connect the iterative solving process of a constrained optimization problem with the temporal evolution of a feedback control system.
Building on this framework, we introduce the PLO method, which leverages MPC as the feedback controller to take predicted constraint violations into consideration. Numerical experiments demonstrated that PLO outperforms the existing PID Lagrangian method regarding feasible regions and average rewards, showcasing its potential for addressing complex constrained RL problems. In the future, we will continue to explore the application of more advanced control methods within this framework.

\bibliographystyle{IEEEtran}


\end{document}